\documentclass{ieeeaccess}

\usepackage{cite}
\usepackage{amsmath,amssymb,amsfonts}
\usepackage{graphicx}
\usepackage{textcomp}
\usepackage{float}
\usepackage{amsmath,amssymb}
\usepackage{upgreek}
\usepackage{booktabs}
\usepackage{multirow}
\usepackage[table,xcdraw]{xcolor}
\usepackage{algorithm, algorithmic}
\usepackage{eqparbox}
\usepackage{etoolbox}
\usepackage{subfigure}

\renewcommand{\thetable}{\Roman{table}}

\def\BibTeX{{\rm B\kern-.05em{\sc i\kern-.025em b}\kern-.08em
    T\kern-.1667em\lower.7ex\hbox{E}\kern-.125emX}}
\begin{document}
\doi{10.48550/arXiv.2309.11427}

\title{Generative Pre-Training of Time-Series Data for Unsupervised Fault Detection in Semiconductor Manufacturing}

\author{\uppercase{Sewoong Lee}\authorrefmark{1}, 
\uppercase{JinKyou Choi}\authorrefmark{1}, and MIN SU KIM\authorrefmark{1}}

\address[1]{Mechatronics Research, Samsung Electronics Co., Ltd., 1-1 Samsungjeonja-ro, Hwaseong-si, Gyeonggi-do 18848, Republic of Korea}

\markboth
{S. Lee \headeretal: Generative Pre-Training of Time-Series Data for Unsupervised Fault Detection in Semiconductor Manufacturing}
{S. Lee \headeretal: Generative Pre-Training of Time-Series Data for Unsupervised Fault Detection in Semiconductor Manufacturing}

\corresp{Corresponding author: Min Su Kim (e-mail: minsu412.kim@samsung.com)}

\begin{abstract}
This paper introduces TRACE-GPT, which stands for \textbf{T}ime-se\textbf{R}ies \textbf{A}nomaly-detection with \textbf{C}onvolutional \textbf{E}mbedding and \textbf{G}enerative \textbf{P}re-trained \textbf{T}ransformers. TRACE-GPT is designed to pre-train univariate time-series sensor data and detect faults on unlabeled datasets in semiconductor manufacturing. In semiconductor industry, classifying abnormal time-series sensor data from normal data is important because it is directly related to wafer defect. However, small, unlabeled, and even mixed training data without enough anomalies make classification tasks difficult. In this research, we capture features of time-series data with temporal convolutional embedding and Generative Pre-trained Transformer (GPT) to classify abnormal sequences from normal sequences using cross entropy loss. We prove that our model shows better performance than previous unsupervised models with both an open dataset, the University of California Riverside (UCR) time-series classification archive, and the process log of our Chemical Vapor Deposition (CVD) equipment. Our model has the highest F1 score at Equal Error Rate (EER) across all datasets and is only 0.026 below the supervised state-of-the-art baseline on the open dataset.
\end{abstract}


\titlepgskip=-21pt

\maketitle

\section{Introduction}
\label{sect:introduction}
\PARstart{E}{VEN} at this moment, even in a leading semiconductor company, equipment engineers are verifying process logs with their own eyes on a daily basis whether it is normal or not. In order to reduce and automate this inefficiency, researches about Fault Detection and Classification (FDC) \cite{lee2017convolutional, fan2020data, park2021artificial, zhu2022cross, brito2022explainable}, time-series classification \cite{xiao2021rtfn, hsu2021multiple, xi2023unsupervised}, or time-series anomaly detection \cite{malhotra2016lstm, audibert2020usad, geiger2020tadgan, wong2022aer} have been conducted.
These research areas are closely interconnected as shown in Fig.~\ref{fig:illustration}.


Despite extensive research efforts, the semiconductor manufacturing industry faces the aforementioned inefficiency due to the following challenges:
\begin{itemize}
  \item First, the lack of abnormal data becomes a challenge. In other words, abnormal data may not have been sufficiently obtained at the time of developing the detection model. The objective of the semiconductor manufacturing industry is to minimize defects, leading to the handling of highly biased datasets with scarce anomalies \cite{hsu2012main}. It is often necessary to develop models that can detect future anomalies even without any abnormal data. In this case, many supervised machine learning methods, such as convolutional neural networks (CNN) \cite{lee2017convolutional} or K-Nearest Neighbor (KNN) \cite{he2007fault}, become inapplicable.
  \item Second, the scarcity of data itself acts as a challenge. Previous studies in time series typically involved training data with sample sizes in the double digits \cite{dau2019ucr}. However, with the accelerated advancements in rapid circuit integration technology, in the semiconductor industry, where development timelines are crucial, it is often necessary to commence model development with only a handful of available data. Given these circumstances, acquiring labeled data becomes even more difficult, making unsupervised learning essential \cite{8913901}. Particularly for precise data labeling, extensive inspection or measurement steps before and after the process are required, leading to increased costs. This conflicts with the fact that fault detection in semiconductor manufacturing is conducted to maximize productivity and profits \cite{verdier2010adaptive}.
  \item Last, in conjunction with the aforementioned challenges, the presence of various mixed types of normal time-series data makes the problem even more complex. For instance, semiconductor manufacturing processes require the continuous adjustment of the process conditions \cite{choi2021deep}. As a result, the definition of a normal time-series can vary. Consequently, datasets can contain mixed types of normality, which makes the use of traditional algorithms such as dynamic time warping (DTW) \cite{bellman1959adaptive} impossible.
\end{itemize}

\Figure[t!](topskip=0pt, botskip=0pt, midskip=0pt)[scale=0.45]{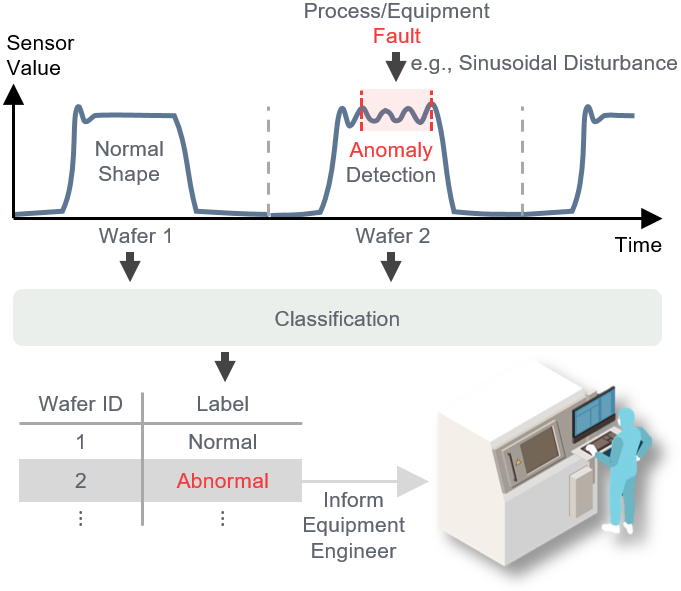} 
{ \textbf{
A illustration of fault detection and classification using time-series anomaly detection in the semiconductor manufacturing industry. 
This illustrates the relationship between fault, anomaly, and abnormal wafer in the semiconductor manufacturing industry, as well as the stages of anomaly detection and classification.
}\label{fig:illustration}}

One of the most commonly used methods in the current semiconductor manufacturing industry involves utilizing CNNs, such as FDC-CNN \cite{lee2017convolutional}, to classify normal and abnormal data through supervised learning. 
They used a combination of feature engineering techniques and supervised learning algorithms to achieve high accuracy in fault detection and classification. 
Even though it demonstrates good performance when applicable, there are many cases in the real world where it cannot be used due to the high cost of building labeled data.
In unsupervised learning, however, even the most recent research \cite{brito2022explainable} attempting unsupervised fault classification on manufacturing data faces the limitation of having to fine-tune the hyperparameters of the anomaly detection model through the fault detection stages, partially using a supervised approach.

Across various fields, including but not limited to the semiconductor manufacturing industry, a prominent domain exploring the characteristics of time-series data through unsupervised learning is time-series representation. 
Recently, TS-TCC \cite{eldele2021time}, TST \cite{zerveas2021transformer}, and TS2Vec \cite{yue2022ts2vec} aimed to develop an unsupervised representation technique for time-series data, which could be used for various tasks such as classification, anomaly detection, and forecasting. Nevertheless, the limitation of unsupervised representation approaches is that they rely on supervised learning methods, such as Support Vector Machines (SVM), when constructing classifiers.

To tackle the challenge of developing unsupervised time-series classification models, the Robust Temporal Feature Network (RTFN) investigates a deep learning approach \cite{xiao2021rtfn}. 
However, RTFN has a limitation in that it can perform unsupervised clustering, but ultimately, in order to extract features for classification, it needs to be embedded into a supervised structure. 
More recently, a model called the Multimodal Domain Adversarial Network (MDAN) has shown promising performance in time-series classification by incorporating both time-domain and frequency-domain feature representations \cite{xi2023unsupervised}. 
However, MDAN has the limitation of requiring a substantial amount of data during the training process for transfer learning from source to target data, and it has been selectively studied only on electrocardiogram or motion sensor data. 
In contrast, our study aims to address issues in semiconductor manufacturing processes where only limited data is provided or in situations where a mix of various sensors such as gas, electric, pressure, and light emission are present.

To achieve end-to-end unsupervised classification as shown in Fig.~\ref{fig:illustration}, unsupervised time-series anomaly detection methods can be utilized. 
Time-series anomaly detection methods are able to find anomalous subsequences of varied lengths within time-series, which can be either single point or collective points \cite{geiger2020tadgan}. Nonetheless, solving real-world problems remains challenging. Our study shows the limitations of existing unsupervised time-series anomaly detection methods on datasets of the semiconductor manufacturing industry.

Humans, despite the challenges mentioned earlier, can relatively easily discern anomalies that existing models struggle with. 
One of the key motivations of this research is that a certain type of deep learning model can detect infinite kinds of potential anomalies of time-series data in semiconductor manufacturing. In other domains such as image processing or Natural Language Processing (NLP), generative pre-training is also proposed to help deep learning with unlabeled data \cite{erhan2009difficulty} and used to improve performance for image processing tasks \cite{erhan2010does}. After the Transformer model is suggested \cite{vaswani2017attention}, the Transformer based model such as Generative Pre-trained Transformers (GPT) \cite{radford2018improving} was developed and outperformed most of state-of-the-art benchmarks in NLP tasks \cite{radford2019language,brown2020language,openai2023gpt4}.

In this paper, we introduce a novel deep learning architecture, TRACE-GPT: \textbf{T}ime-se\textbf{R}ies \textbf{A}nomaly-detection with \textbf{C}onvolutional \textbf{E}mbedding and \textbf{G}enerative \textbf{P}re-trained \textbf{T}ransformers. We demonstrate that the Transformer decoder can be effective in learning sequential features of semiconductor manufacturing sensor data with temporal convolutional networks. The key contributions of this paper are as follows:
\begin{itemize}
  \item {We propose a novel deep learning architecture, TRACE-GPT, for unsupervised time-seres anomaly detection in semiconductor manufacturing. We further developed a model that can classify abnormal data by observing unlabeled data, like a human, without any prior knowledge of anomalies or possible anomalies.}
  \item {We introduce a modified version of the Transformer architecture, originally designed for the natural language processing domain, to effectively handle time-series data. The idea behind the Transformer architecture is to predict the next words in a sentence, but we extended it to learn from numerical sequences and predict the next sensor values.}
  \item{In consideration of real-world challenges encountered in the semiconductor manufacturing industry, we evaluated our model on extreme datasets, including the following scenarios: insufficient abnormal data, single-digit-sized training data, unlabeled data, or mixed normal types. We demonstrated that our model outperforms other baselines.}
\end{itemize}

The structure of this paper is as follows: Section~\ref{sect:related} introduces the related works associated with this study. In Section~\ref{sect:trace}, we describe the architecture and methodology of the deep learning model we propose. In Section~\ref{sect:data}, we provide a detailed explanation of the datasets from the semiconductor manufacturing industry that we conducted experiments on. Section~\ref{sect:expriment} discusses how we set up the experiments and presents the results based on these datasets. Finally, in Section~\ref{sect:conclusion}, we summarize the aspects in which we achieved the motivation of our research.

\section{Related Work}
\label{sect:related}

\subsection{Unsupervised Time-series Anomaly Detection}
\label{sect:utad}

Over the past decades, not to mention the semiconductor manufacturing industry, the rapid proliferation of sensors has increased the demand for temporal observation data, known as time-series data \cite{liu2022mtv}. As it is crucial to detect faults or abnormalities in this type of data, time-series anomaly detection model have shown a great promise \cite{choi2021deep}. 
Time-series anomaly detection involves assigning anomaly scores to identify variable-length anomalous points or segments, $A = \{(t_s, t_e) \mid 1 \le t_s < t_e \le n \}$, within the entire time-series, $X = \{x_1, x_2, x_3, ... , x_n\}$ \cite{alnegheimish2022sintel}. 
Due to the difficulty of obtaining ground truth data for anomalous points or segments within the entire time-series, time-series anomaly detection is primarily being researched using unsupervised methods \cite{wong2022aer}. 
Due to the ability to detect anomalies without labeled training data and the prevalence of models that are easy to reproduce through code, unsupervised time series anomaly detection techniques, as illustrated in Fig. 1, are suitable for the semiconductor manufacturing industry. 
In this section, we discuss some of the well-known unsupervised methods that not only exhibit good performance but are also reproducible through available open-source code.

Since the 1970s, statistical methods such as AutoRegressive Integrated Moving Average (ARIMA) \cite{kinney1978arima} have been investigated for anomaly detection using forecast errors. 
Even though ARIMA is fundamentally a univariate model \cite{choi2021deep}, it is still being used as a powerful baseline model on multivariate tasks \cite{geiger2020tadgan}.

Recently, there has been researches employing deep learning techniques. For instance, Long Short-Term Memory (LSTM) \cite{hochreiter1997long} has demonstrated good performance with sequential data. In the domain of unsupervised learning, Auto-Encoders (AE) represent one of the most typical architectures. This has led to the proposal of models like LSTM-based Encoder-Decoder scheme for Anomaly Detection (EncDec-AD) \cite{malhotra2016lstm}, also known as LSTM-AE \cite{geiger2020tadgan}, which combine LSTM and AE structures. LSTM-AE is capable of detecting anomalies in both short time-series (with lengths as small as 30) and long time-series (with lengths as large as 500). 

In unsupervised learning, Generative Adversarial Networks (GAN)\cite{goodfellow2014generative} also successfully performed many anomaly detection tasks in other domains, such as image processing \cite{schlegl2019f, li2018anomaly}. In the field of time-series analysis, following the design of a time-series representation learning method by \cite{yoon2019time}, TadGAN\cite{geiger2020tadgan} ultimately introduced an end-to-end framework for unsupervised anomaly detection. TadGAN, through the utilization of Wasserstein Loss, has achieved stable convergence for GANs in the context of time-series, and has demonstrated exceptional performance on datasets from fields such as aerospace and information technology.

Lastly, models such as UnSupervised Anomaly Detection for multivariate time series (USAD) \cite{audibert2020usad}, which is inspired by both GAN and AE, have also been proposed. USAD have demonstrated good performance on datasets from fields such as hydrology and environmental monitoring. In our paper, we will examine the limitations of these well-known approaches in terms of their performance on semiconductor manufacturing data.

\subsection{Generative Pre-trained Transformer}
\label{sect:gpt}

\Figure[t!](topskip=0pt, botskip=0pt, midskip=0pt)[scale=0.05]{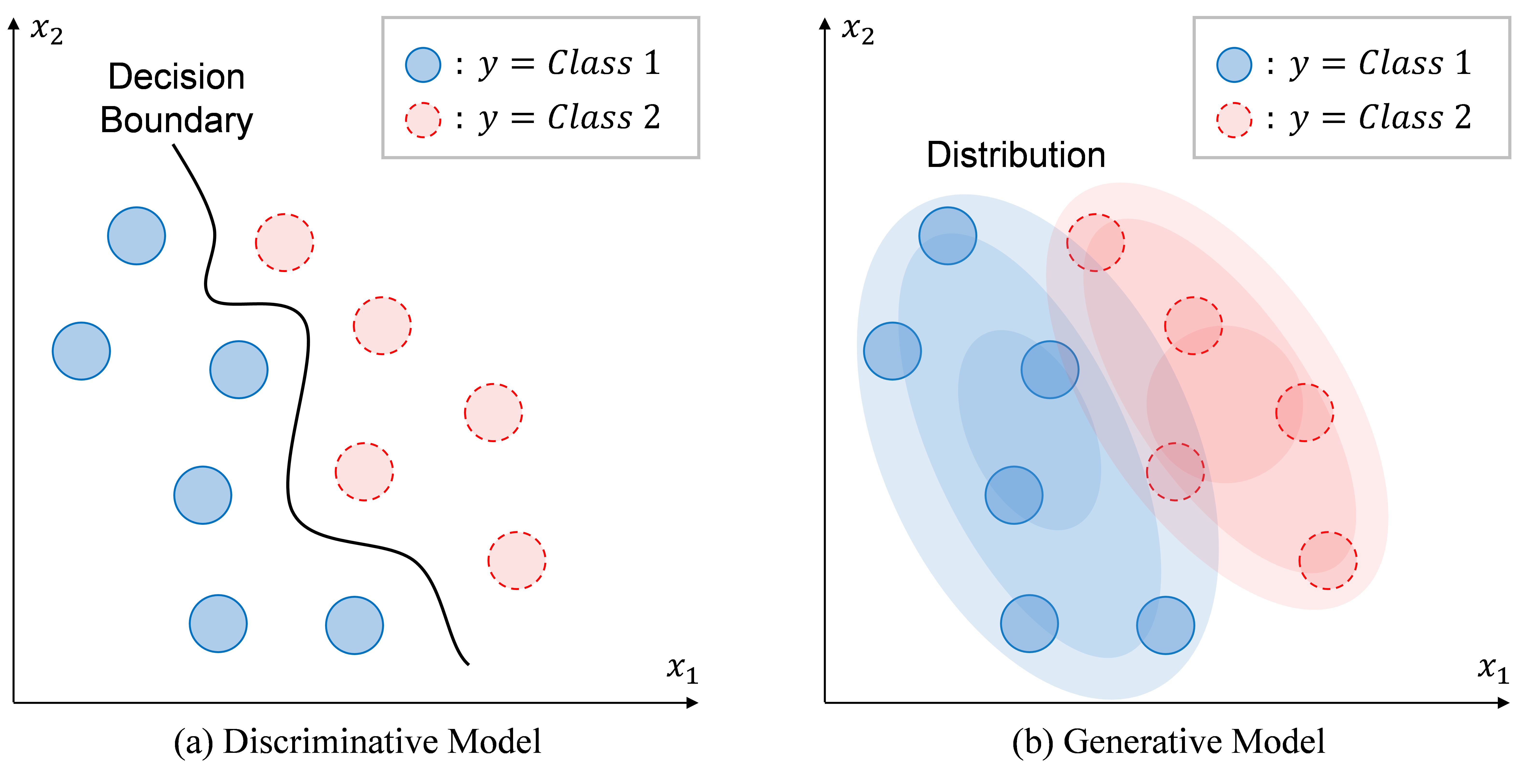} 
{ \textbf{
Comparison between a discriminative model and a generative model. 
The objective of a discriminative model is to learn the decision boundary, whereas the objective of a generative model is to learn the probabilities based on the conditions of features. Generative models ultimately classify labels based on the probabilities obtained in this manner. 
In generative model, visualizing probabilities can enhance the explainability of the model.
}\label{fig:disc_gene}}

As depicted in Fig.~\ref{fig:disc_gene}, in machine learning, generative models learn the distribution of individual classes, whereas discriminative models learn the decision boundary between classes \cite{tu2007learning}. 
Both methods have been explored for sequential data \cite{abou2004classification}. 
Recently, by predicting the probability distribution of the next word in a sentence, generative pre-training has demonstrated significant progress in the field of NLP \cite{radford2019language,brown2020language,openai2023gpt4}. 
In NLP, after unsupervised pre-training stage, pre-trained models typically undergo fine-tuning stage for specific supervised target tasks, such as natural language inference, question answering, or classification. 
Generative pre-training stage is advantageous because it enables the model to learn by predicting the next word, even without labels for specific tasks in the training data. This method was highly effective because the scarcity of labels was a challenge in NLP  \cite{radford2018improving}.

Although the scarcity of labeled training data is one of the major challenges that the semiconductor industry faces as well, the same method cannot be directly applied in this domain. Typically, when processing word tokens, NLP models use the Transformer architecture introduced by [1], which is designed for categorical data such as natural language. Consequently, to process numerical time-series data, the original Transformer architecture needs modification. 
Recently, Anomaly Transformer \cite{xu2021anomaly}, which utilizes the Transformer model as a partial component within the two-branch structure to calculate associations and reconstruct the original sequence, has also been explored. 
However, due to the architectural necessity of reconstructing the entire sequence instead of predicting within periodic units, it has the limitation of requiring a window size of at least a hundred, making it impossible to apply to general semiconductor manufacturing data with short process steps.
Furthermore, rather than employing reconstruction, our work designed the transformer to conduct one-step prediction with task-agnostic generative pre-training.
By visualizing model's probability prediction as shown in Fig.~\ref{fig:disc_gene}, our model offers enhanced explainability.

\Figure[t!](topskip=0pt, botskip=0pt, midskip=0pt)[scale=0.09]{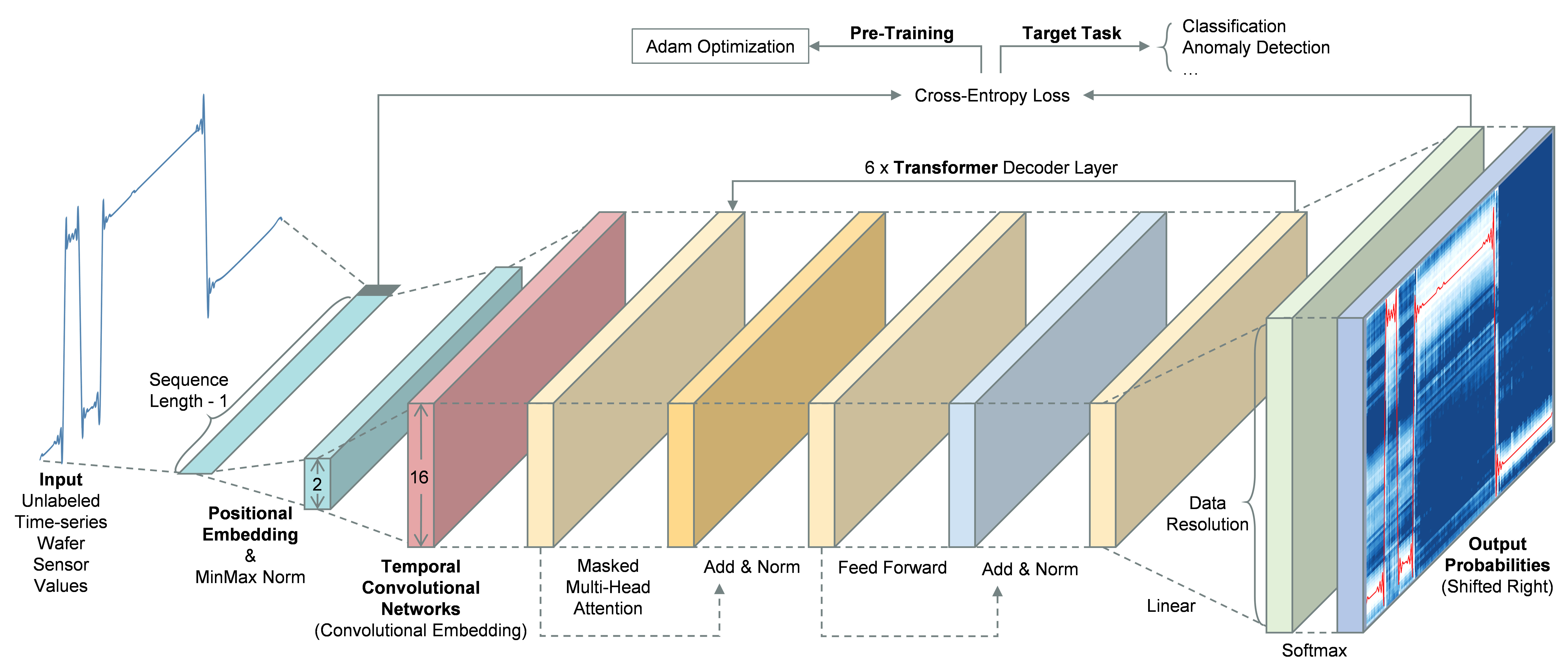} 
{ \textbf{
Deep learning architecture of our model, TRACE-GPT. Even with unlabeled training data, our model can pre-train the temporal features of a time-series sequence, which can ultimately be used in other specific target tasks. As our model predicts the next value in the sequence, output probabilities are shifted right compared to the input sequence.
}\label{fig:architecture}}

\section{TRACE-GPT}
\label{sect:trace}

In this paper, we present a new generative pre-training model for unsupervised time-series anomaly detection and classification. Firstly, as a generative model, our purpose is to learn the underlying probability distribution of the data (See Fig.~\ref{fig:disc_gene}). Secondly, as a pre-training model, our approach allows us to learn the temporal features of data without relying on any specific target task or requiring ground truth labels. 
That is, through one step prediction, the model learns the shapes of the time series by predicting the probability distribution of the next sensor value. By using this approach, each sensor value acts as a label, allowing for effective learning even with unlabeled data that has a small number of sequences. 
Therefore, considering the $n$-th wafer out of a total of N wafers, where each wafer has a time-series sequence $\textbf{X}_{n} = \{x_1, ... , x_T\}$ of length $T$, our objective during pre-training is to maximize the following likelihood.

\renewcommand\algorithmiccomment[1]{%
  \hfill\#\ \eqparbox{COMMENT}{#1}%
}

\begin{equation}
L(\textbf{X}_{n}) = \sum\limits_{i} \log P(x_i|x_1, ... , x_{i-1}; \textbf{w})
\label{mask}
\end{equation}

where $i \in \{2, ... , T\}$, and the conditional probability $P$ is modeled using a neural network with parameters $\textbf{w}$ \cite{radford2018improving}.
As the model is trained to predict the next value based on previous data, ultimately it is structured to receive the preceding $T-1$ values as input and return shifted probabilities to the right, as depicted in the final part of Fig.~\ref{fig:architecture}.

\begin{algorithm}
\caption{Training and Test Phase of TRACE-GPT}\label{algo:tracegpt}
\begin{algorithmic}[1]

\REQUIRE {$ $}
	\newline {$\textbf{X} \in \mathbb{R}^{N \times T} 
	: \textbf{X}_{n, t} \in [0, r)$,
	input data.}
	\newline {$r$, data resolution.}
	\newline {$epoch$, number of iterations over data.}
	\newline {$\eta$, step size.}

\ENSURE
\FOR [Training Phase] {\textbf{each} $epoch$} 
	\FOR {$n = 1, ... , N_{training}$} 
		\FOR {$t = 1, ... , (T-1)$}
			\STATE {$\textbf{X}'_{n,t} = \textbf{X}_{n,t+1}$}
			\STATE {$\textbf{P}_{t} = \mathcal{T}(\mathcal{C}([\textbf{X}_{n,t},t]))$}
		\ENDFOR
		\STATE {$\textbf{g}_{\textbf{w}_{\mathcal{C}, \mathcal{T}}}
		= \nabla (CrossEntropy(
		\textbf{P}^{\intercal}, 
		\lfloor\textbf{X}'_{n}\rfloor))$}
		\STATE {$\textbf{w}_{\mathcal{C}, \mathcal{T}}
		= \textbf{w}_{\mathcal{C}, \mathcal{T}} 
		+ \eta\cdot Adam(
		\textbf{w}_{\mathcal{C}, \mathcal{T}}, 
		\textbf{g}_{\textbf{w}_{\mathcal{C}, \mathcal{T}}})$}
	\ENDFOR
\ENDFOR
\FOR [Test Phase] {$n = N_{training+1}, ... , N$} 
	\FOR {$t = 1, ... , (T-1)$}
		\STATE {$\textbf{X}'_{n,t} = \textbf{X}_{n,t+1}$}
		\STATE {$\textbf{P}_{t} = \mathcal{T}(\mathcal{C}([\textbf{X}_{n,t},t]))$}
	\ENDFOR
	\STATE $\mathcal{L}_{n} = CrossEntropy(
	\textbf{P}^{\intercal},
	\lfloor\textbf{X}'_{n}\rfloor))$
	\STATE $output\_probabilities_{n} = Softmax(\textbf{P}, \texttt{dim}=1)$
\ENDFOR
\end{algorithmic}
\label{algo}
\end{algorithm}

\Figure[t!](topskip=0pt, botskip=0pt, midskip=0pt)[scale=0.125]{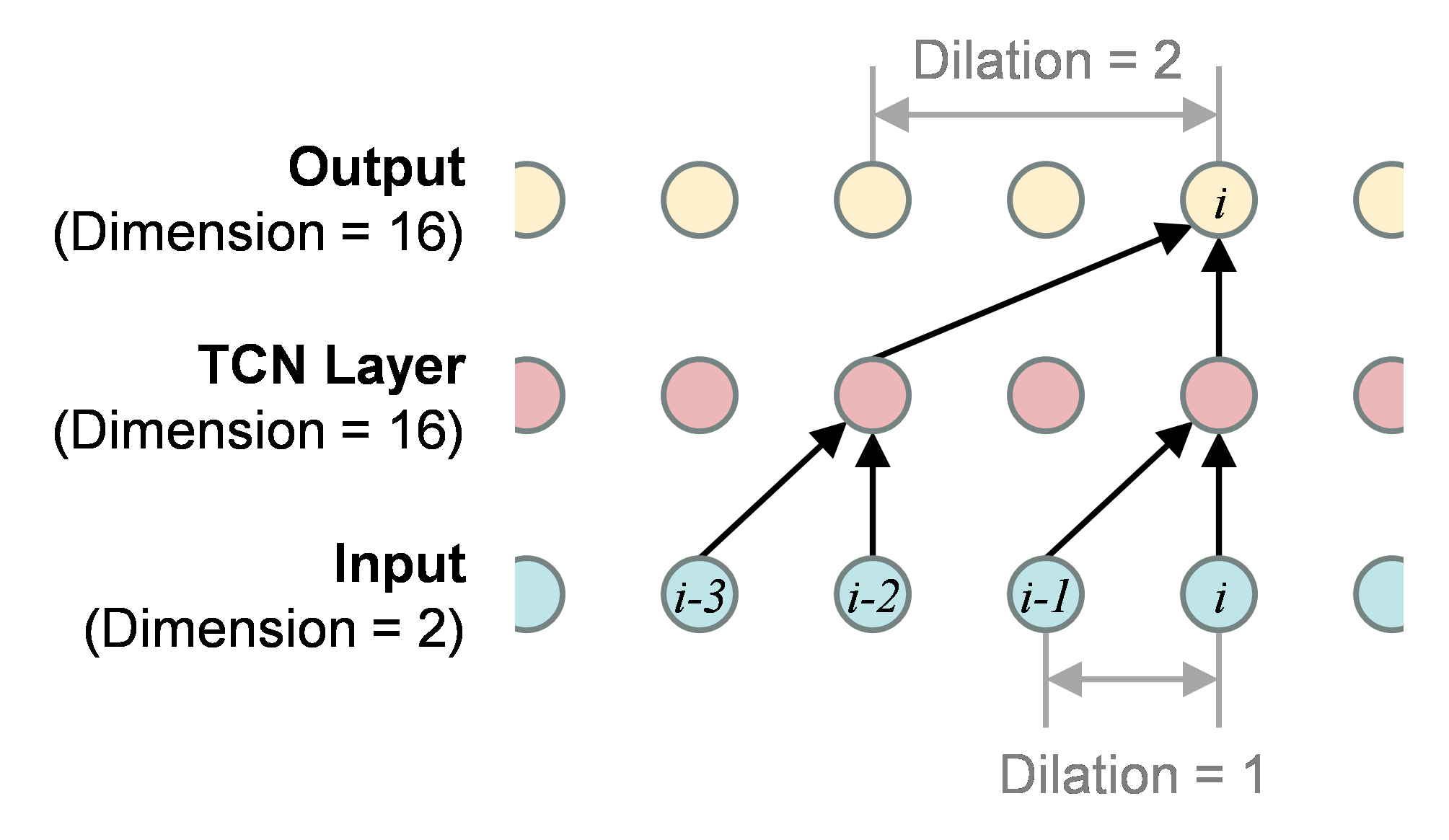}{\textbf{
Visualization of the convolutional embedding process using a Temporal Convolutional Network (TCN). 
TCN is similar to CNN, but it differs in that future data are masked, and only past data are convolved with the filter for computation. This distinction makes it useful in deep learning architectures for predicting future data. Additionally, when multiple layers are employed, the receptive field that captures past data expands as dilation increases. These characteristics enable the model to effectively learn and utilize fluctuations in sensor values prior to the prediction point.
}\label{fig:tcn}}

\subsection{Positional Embedding (PE)}
As a first step, we want our model to be aware of the global position within the whole sequence. 
To prevent the loss of sequential information when features are computed in machine learning or neural networks, a methodology that incorporates temporal feature information is referred to as positional embedding. 
There can be several ways to embed position, but a simple approach is to normalize and inject it as another dimension, as $\textbf{X}_{n,t} \rightarrow [\textbf{X}_{n,t},t]$. This is suitable when the sequence length is fixed, and the model does not need to extrapolate to sequence lengths \cite{vaswani2017attention}. 

For semiconductor manufacturing data, in typical cases that do not use end point detection, it is common to maintain a consistent process time to ensure identical process outcomes \cite{lin1999effective}. In the NLP domain, sentence lengths can vary widely, and in the finance or hydrology fields, the length of accumulating time series data can become indefinitely long. In contrast, the semiconductor manufacturing industry is characterized by periodicity at the wafer level, and maintaining consistency is crucial, highlighting these distinct advantages. Therefore, our model used positional embedding by injecting a positional index and stacking arrays. 
To mitigate the influence of varying scales between positional index and sensor values, which could lead to unequal feature influences and delayed learning, we applied min-max normalization alongside.

\subsection{Temporal Convolutional Networks (TCN)}
In other domains like NLP, an embedding layer such as word2vec \cite{karani2018introduction} is commonly used to convert categorical data into high dimensional numerical data. 
How can we extract enough temporal features with univariate semiconductor manufacturing sensor data? One approach suggested by recent research is using Temporal Convolutional Networks (TCN) \cite{kalchbrenner2016neural} which belong to the class of CNN designed for processing sequential data and capturing local temporal changes \cite{lea2017temporal}.

The faults in the semiconductor manufacturing industry easily retain temporal characteristics in time-series data that are suitable for embedding using TCN. 
For example, anomalies like sudden spikes, drops, or peripheral points introduce discrepancies with previously recorded values. 
As a result, when predicting the normal shape using convolution results from past data, differences arise at abnormal points. 
Therefore, TCN becomes an effective approach for embedding into multiple dimensions through 1D convolutions. 
For this reasons, our model employs TCN as one of the steps for embedding dimensions, as shown in Fig.~\ref{fig:architecture}.
Empirically, we used a hidden TCN layer, as shown in Fig.~\ref{fig:tcn}, and the dimension $d$ is set to 16. We denote this process of convolutional embedding as $\mathcal{C}: \mathbb{R}^{2} \rightarrow \mathbb{R}^{d}$.

\subsection{Transformer}
The Transformer has demonstrated strong performance across various NLP tasks \cite{radford2018improving} and offers several advantages when processing sequential data. 
Firstly, its attention mechanism allows for parallel processing, resulting in better time performance compared to models like LSTM \cite{vaswani2017attention}. 
Among the entire Transformer architecture, the Transformer Decoder Layer is particularly well-suited for semiconductor manufacturing data, where the values in the later portion are masked to predict the next values. This characteristic enables the model to sequentially assess whether the next value deviates from the expected, thus facilitating the detection of anomalies, similar to how equipment engineers analyze them.

In line 5 of Algorithm~\ref{algo}, the first transformation of the Transformer decoder layer $\mathcal{T}: \mathbb{R}^{d} \rightarrow \mathbb{R}^{r}$ is $Q_{i}, K_{i},$ and $V_{i}$, which are linear transformation of the $(T-1) \times d$ size input of the i-th attention head as \eqref{attention}, representing the query, key, and value, respectively. 
As the original Transformer research \cite{vaswani2017attention}, we employ $h = 8$ parallel attention heads. With the Transformer decoder layer, masked multi-head attention can transform the sequence of the previous values in order to predict next values with attention scores.

\begin{equation}
Attention_{i}(Q_{i},K_{i},V_{i}) 
= Softmax(\frac{Q_{i}K_{i}^{\intercal}}{\sqrt{d_{k}}}+M) \cdot V_{i}
\label{attention}
\end{equation}

where $M$ is the mask of $\mathbb{R}^{(T-1) \times (T-1)}$ with values defined by \eqref{mask}, designed to prevent the model from peeking at the next value when making predictions.

\begin{equation}
M_{i,j} = 
\left\{ 
  \begin{array}{ c l }
    -\infty		& \quad \textrm{if} ~~i < j, \\
    0			& \quad \textrm{otherwise.}
  \end{array}
\right.
\label{mask}
\end{equation}

When denoting Masked Multi-Head Attention as $MMHA$, $MMHA$ can be expressed as shown in \eqref{mmha}, which is subsequently transformed by a Feed Forward Network ($FFN$) as in \eqref{ffn} with Rectified Linear Unit ($ReLU$) and layer normalization \cite{ba2016layer}.

\begin{equation}
MMHA(Q,K,V) 
= Concat^{h}_{i=1}\{Attention_{i}(Q_{i},K_{i},V_{i})\}
\label{mmha}
\end{equation}

\begin{equation}
FFN(x) 
= ReLU(xW_1 + b_1) \cdot W_2 + b_2
\label{ffn}
\end{equation}

Finally, after six iterations, as shown in Fig.~\ref{fig:architecture}, the linear transformation of values based on the data resolution is applied after the Transformer decoder layers. 
Generally, data resolution refers to the level of detail or granularity and represents the smallest difference between adjacent data points in terms of value, time, or distance.
In this paper, we use the term `data resolution', $r \in \mathbb{N}$, to refer to the number of value levels that the model can distinguish.
In other words, if a model has a data resolution of 100, it means that all values are mapped to integers within the range from 0 to 100, where 0 represents the minimum value and 100 represents the maximum value.
In this paper, based on the observation that humans tend to classify anomalies in sensor data using percentage units, we employed a resolution of 100 in our model. As a result, by iterating lines 3-8 of Algorithm~\ref{algo}, we can obtain the shifted right probability $\textbf{P}_t \in \mathbb{R}^{r}$ and pre-trained weights $\textbf{w}_{\mathcal{C}, \mathcal{T}}$ that are irrelevant to the specific task.

\subsection{Applications of Our Model}

The difference of generative pre-trained models in time-series compared to NLP lies in the fact that the trained model does not undergo supervised training to perform the target task. In NLP, the sequences generated by the model can serve as labels that can be used for the target tasks, such as translation or question answering. However, in time-series, the cross entropy between the possibilities generated by the model and the original sequences can be used for the target task as in lines 11-18 of Algorithm~\ref{algo}.
As shown in Fig.~\ref{fig:architecture}, by calculating the cross-entropy loss $\mathcal{L}$, a typical loss function used between a sequence and multi-class probabilities, and minimizing it with $Adam$ \cite{kingma2014adam}, the most commonly used optimizer, the model can learn how to predict next value similar to the actual value. 
As in line 17 of Algorithm~\ref{algo}, by passing through the softmax function, the model can visualize the results in an explainable graph, as shown in Fig.~\ref{fig:vis_all} (c) and (d), so that equipment engineers can comprehend the results.

Our specific task can be either calculating anomaly score or classifying abnormal sequences, as shown in Fig.~\ref{fig:illustration}.
For anomaly detection, $\mathcal{L}_{n, t}$ can be used as anomaly score at timpstamp $t$.
For classification, cross-entropy loss of $n$-th sequence $\mathcal{L}_n=\sum\limits_t{\mathcal{L}_{n, t}}$ can be used to measure how much the predicted value significantly deviates from the actual value. Based on training loss distribution, as depicted in Fig.~\ref{fig:cvd_hist}, threshold $\uptau$ can be set and utilized to classify abnormal sequences from normal sequences.

\begin{equation}
Classifier(\textbf{X}_n) = 
\left\{ 
  \begin{array}{ c l }
    normal		& \quad \textrm{if} ~~\uptau > \sum\limits_t{\mathcal{L}_{n, t}}, \\
    abnormal	& \quad \textrm{otherwise.}
  \end{array}
\right.
\label{eq:classification}
\end{equation}

\section{Data}
\label{sect:data}

\begin{table}
\caption{\textbf{The characteristics and size of the datasets.}}
\label{tab:datasets}
\begin{tabular}{p{40pt}p{50pt}p{50pt}p{50pt}}
\toprule
                          	&          				& CVD 						& UCR\\ \toprule
\multicolumn{2}{l}{Challenges}						& Small size				& Mixed Types\\
\multicolumn{2}{l}{}								& No abnormal$^{\mathrm{a}}$& Unlabeled$^{\mathrm{b}}$\\ \midrule
\multicolumn{2}{l}{Sequence Length / Wafer}			& 53  						& 152\\ \midrule
\multirow{3}{*}{Training}	& Normal   				& 9   						& 903\\
                        	& Abnormal 				& -							& 97\\ \cmidrule(l){2-4} 
                          	& Total ($N_{training}$)& 9   						& 1000\\ \midrule
\multirow{3}{*}{Test}     	& Normal   				& 567 						& 5499\\
                          	& Abnormal 				& 8   						& 665\\ \cmidrule(l){2-4} 
                          	& Total ($N_{test}$)   	& 575 						& 6164\\ \bottomrule
\multicolumn{4}{l}{$^{\mathrm{a}}$Abnormal data are not available in training phase.}\\
\multicolumn{4}{l}{$^{\mathrm{b}}$Normal or abnormal labels are not used in training phase.}\\
\end{tabular}
\end{table}

In the cost-sensitive semiconductor manufacturing industry, datasets with labeling at each time stamp, as typically used in time series anomaly detection, are generally not employed due to labeling expenses \cite{8913901, verdier2010adaptive}. 
Instead, the assessment of defects is commonly done at the wafer level through measurements or inspections. Therefore, using datasets where normal and anomaly classifications are assigned at the wafer's sequence level, as depicted in Fig.~\ref{fig:illustration}, allows for the evaluation of classification performance. For these experiments, as shown in Table~\ref{tab:datasets}, we experimented with the challenges of the semiconductor manufacturing industry using two individual univariate datasets, including one open dataset\footnote{https://www.timeseriesclassification.com/description.php?Dataset=Wafer}.

\subsection{CVD Equipment Process Log}
\label{sect:cvd}

We use the process log of the deposition step from one of our Chemical Vapor Deposition (CVD) equipment testbeds. The data consist of Mass Flow Controller (MFC) sensor data, which is used to measure a gas flow rate. Since the length of the deposition step time is controlled to be equal, the lengths of the sequences are also equal as Table~\ref{tab:datasets}. Whether it is normal or abnormal has been confirmed by the engineer in charge. The training data are from the first nine wafers since this equipment was installed, and the normal test data are 567 wafers processed thereafter.

\Figure[t!](topskip=0pt, botskip=0pt, midskip=0pt)[scale=0.5]{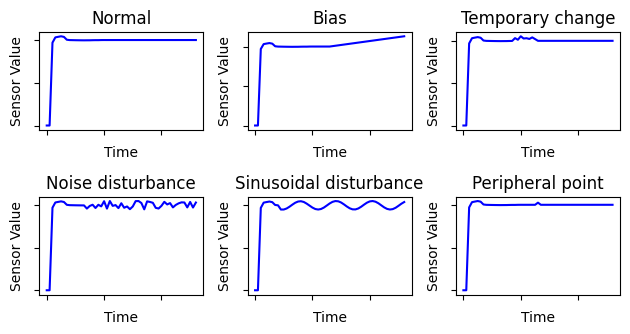}{\textbf{
Compared to the normal data in the upper left, the illustration depicts the five fault types proposed by \cite{mellah2022semiconductor}. 
The five abnormal behaviors classified as fault types occur within a maximum range of 5\% from the normal values.
}\label{fig:fault_five}}

\Figure[t!](topskip=0pt, botskip=0pt, midskip=0pt)[scale=0.5]{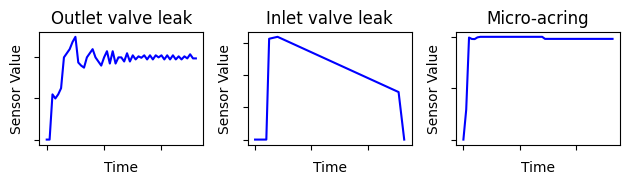}{\textbf{
The illustration depicts the three fault types that occurred in our test vehicle. From left to right, when an outlet valve leak occurred, it exhibited a damped oscillation pattern; when an inlet valve leak occurred, the set point was not reached and sensor values gradually decreased; and when micro-arcing occurred, the sensor value showed a drop of approximately 1.5\% around the mid-point of the process.
}\label{fig:fault_ours}}

Abnormal test data are obtained or generated in the following ways. Firstly, there were two abnormal cases that occurred in our test vehicle during the data collection period: inlet valve leak and outlet valve leak. Secondly, we generated a sequence to detect micro-arcing phenomena that may occur within the semiconductor manufacturing chamber.
In reference to previous research \cite{yan2009solving} that it can be detected through fluctuations in electric current sensor values and the values that actually occured in one of our test vehicles, we added a sequence that reduced the data value by 1.5\% from the midpoint.
Lastly, we added five fault types in the previous suggested semiconductor time-series anomaly classification research \cite{mellah2022semiconductor}: bias, temporary change, noise disturbance, sinusoidal disturbance, and peripheral point.
For example, in the case of noise disturbance, it is a commonly measurable fault type in motion sensors \cite{hoo2019biometric}.
Although the CVD dataset is based on gas flow data, these various fault types were used to examine the generality and robustness of the model for manufacturing data.

In summary, we conducted tests on the scenario with a fault rate of 1.39\%, which was augmented based on previous research \cite{mellah2022semiconductor}, using data with a fault rate of 0.35\% relative to a total of 567 normal wafer samples.
The dataset is based on data acquired from real semiconductor manufacturing equipment with extremely low fault occurrence frequency. Therefore, proper detection can lead to swift improvements in equipment or process issues.
Particularly, faults such as micro-arcing, temporary change, and peripheral point are types that cannot be detected by the conventional Lower Specification Limits (LSL) and Upper Specification Limits (USL) interlocks commonly employed in the current equipment \cite{hofer2017determination}. As a result, this provides an effect of reducing temporal losses.
If real-time detection is achievable through a deep learning model for this dataset, it would have a significant impact on improving the yield of hundreds of wafers lost in the process of reaching the measurement stage, confirming the issue, and subsequently enhancing the equipment or process, based on the current CVD equipment.
Since it is a dataset where fault types are clearly identified, we could conduct experiments to see which types of abnormalities are well classified and which types are difficult to classify, as shown in Fig.~\ref{fig:cvd_hist}.

\subsection{The UCR Time-Series Classification Archive}

To the best of our knowledge, the wafer sensor data in the University of California Riverside (UCR) time-series classification archive \cite{dau2019ucr} are virtually the only open dataset available for time-series sensor data in semiconductor manufacturing \cite{espadinha2021review}, which consists of six sensor values used in the etching process: radio frequency forward power, radio frequency reflected power, chamber pressure, 405 nanometer emission, 520 nanometer emission, and direct current bias \cite{olszewski2001generalized}. Generally, the UCR datasets are used to compare universal time-series classification algorithms \cite{batista2014cid, wu2021current}. We selected this dataset to prove that our model can outperform other unsupervised models on wafer sensor data and even be comparable to state-of-the-art supervised models \cite{guillaume2022random}.

It is notable that the six sensor values are combined in a single dataset, requiring the model to be robust in processing different types of sensors simultaneously. This indicates the importance of developing a model that can effectively handle mixed sensor data and maintain its performance regardless of the sensor type. There are total 7,164 sequences with two classes, a normal and an abnormal. 10.6\% of the total data are abnormal. The training and test data have been pre-divided by the original paper, with 1000 sequences specifically allocated for the training set.

With supervised learning, the current state-of-the-art model has achieved almost perfect classification with accuracy of 0.99995 \cite{guillaume2022random}. Our work, however, aims to solve the challenge of training without labeled data, with unsupervised learning. For unsupervised learning, all experiments were conducted without the labels of the training data. In other words, during the training phase, all labels were removed, and information about the classes or their corresponding labels was not used.

\begin{table*}[t]
\caption{Accuracy, Precision, Recall, and F1 score at EER for baseline models. From red to blue, evaluation metrics increase from 0 to 1.}
\label{tab:baselines}
\resizebox{\linewidth}{!}{%

\begin{tabular}{l|cccc|cccc|cccc}
\hline
\multicolumn{1}{c|}{} &
  \multicolumn{4}{c|}{CVD} &
  \multicolumn{4}{c|}{UCR} &
  \multicolumn{4}{c}{Average} \\ \cline{2-13} 
\multicolumn{1}{c|}{\multirow{-2}{*}{Models}} &
  Accuracy &
  Precision &
  Recall &
  F1 &
  Accuracy &
  Precision &
  Recall &
  F1 &
  Accuracy &
  Precision &
  Recall &
  F1 \\ \cline{1-1}
TRACE-GPT &
  \cellcolor[HTML]{5A8AC6}1.000 &
  \cellcolor[HTML]{5A8AC6}1.000 &
  \cellcolor[HTML]{5A8AC6}1.000 &
  \cellcolor[HTML]{5A8AC6}1.000 &
  \cellcolor[HTML]{A5BFE1}0.955 &
  \cellcolor[HTML]{6391CA}0.995 &
  \cellcolor[HTML]{A5BFE1}0.954 &
  \cellcolor[HTML]{85A8D5}0.974 &
  \cellcolor[HTML]{80A5D4}0.977 &
  \cellcolor[HTML]{5F8EC8}0.997 &
  \cellcolor[HTML]{80A5D4}0.977 &
  \cellcolor[HTML]{7099CE}0.987 \\
ARIMA &
  \cellcolor[HTML]{6E98CD}0.988 &
  \cellcolor[HTML]{FABFC2}0.538 &
  \cellcolor[HTML]{FBF7FA}0.875 &
  \cellcolor[HTML]{FAD4D7}0.667 &
  \cellcolor[HTML]{FBE5E7}0.763 &
  \cellcolor[HTML]{95B4DB}0.964 &
  \cellcolor[HTML]{FBE5E7}0.763 &
  \cellcolor[HTML]{FBF3F6}0.852 &
  \cellcolor[HTML]{FBF7FA}0.876 &
  \cellcolor[HTML]{FBE3E5}0.751 &
  \cellcolor[HTML]{FBEEF1}0.819 &
  \cellcolor[HTML]{FBE4E7}0.759 \\
TadGAN &
  \cellcolor[HTML]{5A8AC6}1.000 &
  \cellcolor[HTML]{5A8AC6}1.000 &
  \cellcolor[HTML]{5A8AC6}1.000 &
  \cellcolor[HTML]{5A8AC6}1.000 &
  \cellcolor[HTML]{FAB8BA}0.495 &
  \cellcolor[HTML]{F87779}0.106 &
  \cellcolor[HTML]{FAB8BB}0.496 &
  \cellcolor[HTML]{F88285}0.175 &
  \cellcolor[HTML]{FBE2E5}0.748 &
  \cellcolor[HTML]{FAC2C4}0.553 &
  \cellcolor[HTML]{FBE2E5}0.748 &
  \cellcolor[HTML]{FAC7CA}0.588 \\
LSTM-AE &
  \cellcolor[HTML]{FAC2C4}0.555 &
  \cellcolor[HTML]{F8696B}0.019 &
  \cellcolor[HTML]{FACDD0}0.625 &
  \cellcolor[HTML]{F86C6E}0.038 &
  \cellcolor[HTML]{FBE5E7}0.764 &
  \cellcolor[HTML]{95B4DB}0.964 &
  \cellcolor[HTML]{FBE5E7}0.764 &
  \cellcolor[HTML]{FBF3F6}0.852 &
  \cellcolor[HTML]{FAD3D6}0.659 &
  \cellcolor[HTML]{FAB7BA}0.492 &
  \cellcolor[HTML]{FBD9DC}0.694 &
  \cellcolor[HTML]{F9AFB2}0.445 \\
USAD &
  \cellcolor[HTML]{FBF0F2}0.829 &
  \cellcolor[HTML]{F86F71}0.058 &
  \cellcolor[HTML]{FBF4F7}0.857 &
  \cellcolor[HTML]{F8777A}0.109 &
  \cellcolor[HTML]{FAC2C5}0.558 &
  \cellcolor[HTML]{E9EFF9}0.913 &
  \cellcolor[HTML]{FAC2C5}0.558 &
  \cellcolor[HTML]{FBD9DC}0.693 &
  \cellcolor[HTML]{FBD9DC}0.694 &
  \cellcolor[HTML]{FAB6B9}0.485 &
  \cellcolor[HTML]{FBDBDE}0.708 &
  \cellcolor[HTML]{F9A8AB}0.401 \\ \hline
\end{tabular}%

}
\end{table*}

\begin{table*}[t]
\caption{Accuracy, Precision, Recall, and F1 score at EER for ablation study. From red to blue, evaluation metrics increase from 0 to 1.}
\label{tab:ablation}
\resizebox{\linewidth}{!}{%

\begin{tabular}{l|cccc|cccc|cccc}
\hline
\multicolumn{1}{c|}{} &
  \multicolumn{4}{c|}{CVD} &
  \multicolumn{4}{c|}{UCR} &
  \multicolumn{4}{c}{Average} \\ \cline{2-13} 
\multicolumn{1}{c|}{\multirow{-2}{*}{Models}} &
  Accuracy &
  Precision &
  Recall &
  F1 &
  Accuracy &
  Precision &
  Recall &
  F1 &
  Accuracy &
  Precision &
  Recall &
  F1 \\ \cline{1-1}
TRACE-GPT &
  \cellcolor[HTML]{5A8AC6}1.000 &
  \cellcolor[HTML]{5A8AC6}1.000 &
  \cellcolor[HTML]{5A8AC6}1.000 &
  \cellcolor[HTML]{5A8AC6}1.000 &
  \cellcolor[HTML]{A9C2E2}0.955 &
  \cellcolor[HTML]{6491CA}0.995 &
  \cellcolor[HTML]{AAC2E2}0.954 &
  \cellcolor[HTML]{88AAD6}0.974 &
  \cellcolor[HTML]{82A6D4}0.977 &
  \cellcolor[HTML]{5F8EC8}0.997 &
  \cellcolor[HTML]{82A6D4}0.977 &
  \cellcolor[HTML]{719ACE}0.987 \\
TRACE-GPT   w/o PE &
  \cellcolor[HTML]{618FC9}0.997 &
  \cellcolor[HTML]{FBEAED}0.800 &
  \cellcolor[HTML]{5A8AC6}1.000 &
  \cellcolor[HTML]{FBF9FC}0.889 &
  \cellcolor[HTML]{FBFBFE}0.901 &
  \cellcolor[HTML]{719BCF}0.987 &
  \cellcolor[HTML]{FBFBFE}0.901 &
  \cellcolor[HTML]{BFD2EA}0.942 &
  \cellcolor[HTML]{B4C9E6}0.949 &
  \cellcolor[HTML]{FBF9FC}0.893 &
  \cellcolor[HTML]{B1C7E5}0.950 &
  \cellcolor[HTML]{EDF2FA}0.915 \\
TRACE-GPT   w/o TCN &
  \cellcolor[HTML]{6D97CD}0.990 &
  \cellcolor[HTML]{FAC4C7}0.571 &
  \cellcolor[HTML]{5A8AC6}1.000 &
  \cellcolor[HTML]{FBDEE1}0.727 &
  \cellcolor[HTML]{C1D3EB}0.941 &
  \cellcolor[HTML]{6894CB}0.993 &
  \cellcolor[HTML]{C2D3EB}0.941 &
  \cellcolor[HTML]{96B4DB}0.966 &
  \cellcolor[HTML]{97B5DC}0.965 &
  \cellcolor[HTML]{FBE7EA}0.782 &
  \cellcolor[HTML]{8EAFD9}0.970 &
  \cellcolor[HTML]{FBF2F4}0.847 \\
TRACE-GPT   w/o Transformer &
  \cellcolor[HTML]{91B1DA}0.969 &
  \cellcolor[HTML]{F9989B}0.308 &
  \cellcolor[HTML]{5A8AC6}1.000 &
  \cellcolor[HTML]{FAB3B6}0.471 &
  \cellcolor[HTML]{FBF7FA}0.881 &
  \cellcolor[HTML]{779ED0}0.984 &
  \cellcolor[HTML]{FBF7FA}0.882 &
  \cellcolor[HTML]{D4E0F1}0.930 &
  \cellcolor[HTML]{DDE6F4}0.925 &
  \cellcolor[HTML]{FAD0D3}0.646 &
  \cellcolor[HTML]{C1D3EB}0.941 &
  \cellcolor[HTML]{FBD9DC}0.700 \\ \hline
\end{tabular}%

}
\end{table*}

\begin{figure*}[t!]
 \centering
 \subfigure[CVD Baselines] {\includegraphics[scale=0.3]{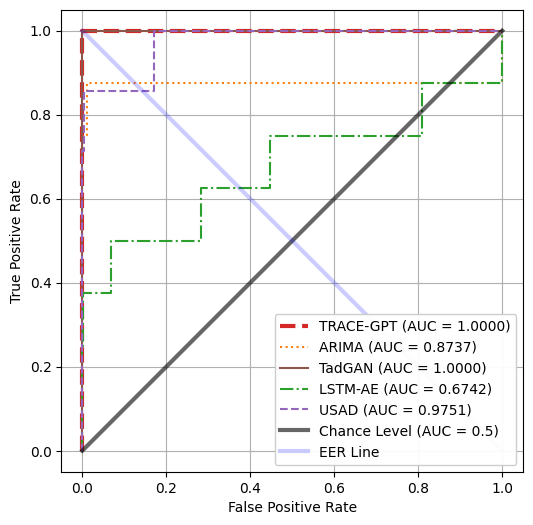}}\hfill
 \subfigure[CVD Ablation] {\includegraphics[scale=0.3]{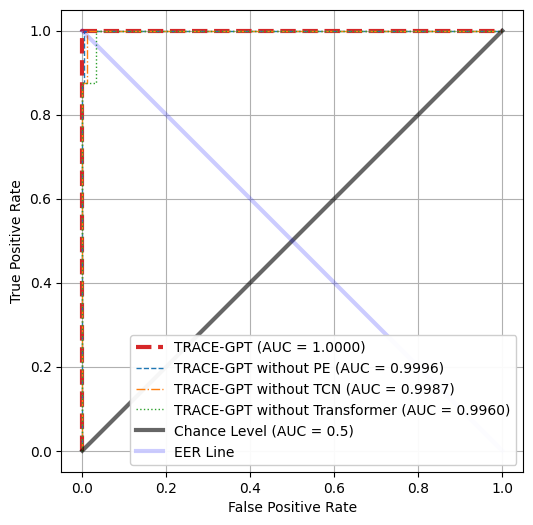}}\hfill
 \subfigure[UCR Baselines] {\includegraphics[scale=0.3]{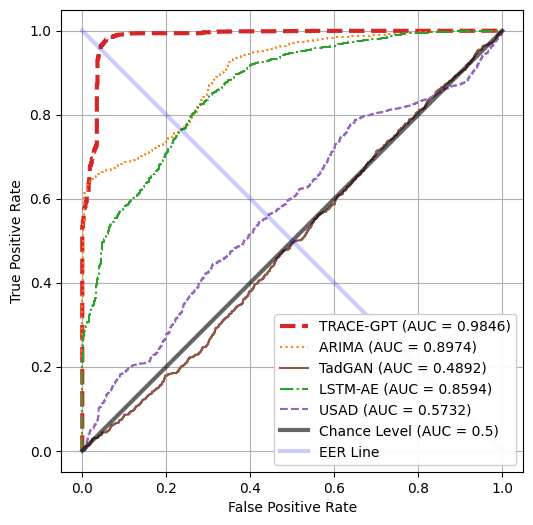}}\hfill
 \subfigure[UCR Ablation] {\includegraphics[scale=0.3]{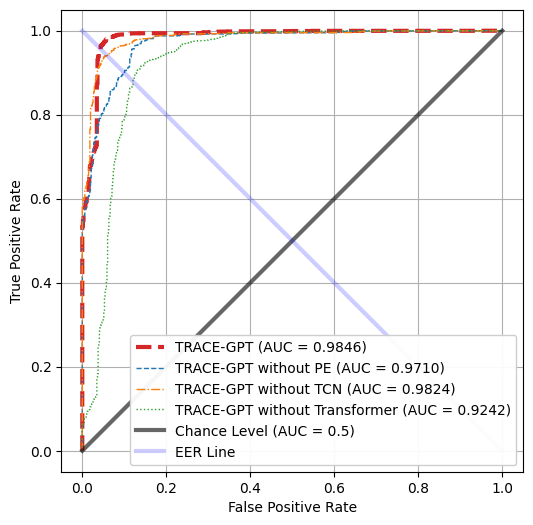}}
 \caption{
 Resulting ROC curves of the proposed TRACE-GPT model. Corresponding AUC values are given in the legend.
 }
 \label{fig:roc}
\end{figure*}

\section{Experimental Results}
\label{sect:expriment}

\subsection{Experimental Setup}

Firstly, all data were normalized to the range [0, 1] based on the minimum and maximum values of the training data. For a total of $N$ wafers with a fixed sequence length of $T$, the entire time-series was concatenated as shown in Fig.~\ref{fig:illustration}, resulting in $\{x_1, ... , x_{{N}\times{T}}\}$.

We measured the performance of different methods using the commonly used metrics such as Accuracy, Precision, Recall, and F1-Score as shown in Table~\ref{tab:baselines} and Table~\ref{tab:ablation}. Specifically, as those metrics of unsupervised learning can vary with the anomaly score threshold, they are measured at the Equal Error Rate (EER) of the Receiver Operating Characteristic (ROC) curve, as shown in Fig.~\ref{fig:roc}. 
To compare performance across all thresholds, the Area Under the ROC Curve (AUC), also known as AUROC, can be analyzed, as shown in Fig.~\ref{fig:roc}.

In the semiconductor manufacturing industry, as depicted in Fig.~\ref{fig:illustration}, the ultimate purpose is to classify whether a sequence of a wafer is normal or abnormal. 
One of the simplest approaches is using the criterion that anomalies are correctly detected when they appear in the abnormal sequence \cite{hundman2018detecting}. However, this approach is susceptible to the change of the anomaly score's threshold. The method to maximize the performance of the baseline model involves converting a collection of anomaly scores into errors and then plotting the ROC curve based on the threshold.
In this regard, \cite{geiger2020tadgan} proposed methods utilizing point-wise difference, area difference, and DTW. We further extended the evaluation metrics by incorporating additional metrics, including Mean Squared Error (MSE) and Mean Squared Logarithmic Error (MSLE), to apply the most effective performance evaluation metric for each baseline model.

We conducted all experiments in one machine with an Intel Xeon W-2125 processor containing 4 CPU cores, and one Nvidia Quadro P2000 GPU containing 1024 CUDA cores and 5 GB GDDR5 on-board memory.

\subsection{Baseline Models}

We include the following baselines in our experiments:

\textbf{ARIMA} \cite{kinney1978arima} is implemented with the \texttt{statsmodels} library. 
We used p=1, d=0, q=0 with MSLE, which was empirically determined as the best setting.

\textbf{TadGAN} \cite{geiger2020tadgan}
is implemented using Orion \cite{alnegheimish2022orion}. 
For our dataset, using point-wise prediction errors showed better performance than DTW. We employed a window size of $\frac{T}{3}$, as specified in the original paper .

\textbf{LSTM-AE} \cite{malhotra2016lstm} uses two LSTM layers, each with 64 units for the encoder and decoder. A pointwise reconstruction error is used to detect anomalies \cite{geiger2020tadgan}.

\textbf{USAD} \cite{audibert2020usad} consists of one encoder and two decoders, ultimately  combining two autoencoders. The encoder and decoders pass through three layers, halving or doubling the dimensions, until reaching the dimensionality $m$ of the latent space. The window size was set to the default setting of 12, and the dimensionality of the latent space was set to 10. According to the original research, further increasing the latent space did not significantly improve performance, and experimentation with higher dimensions was limited by our machine's memory constraints.

\subsection{Benchmarking Results}

\Figure[t!](topskip=0pt, botskip=0pt, midskip=0pt)[scale=0.55]{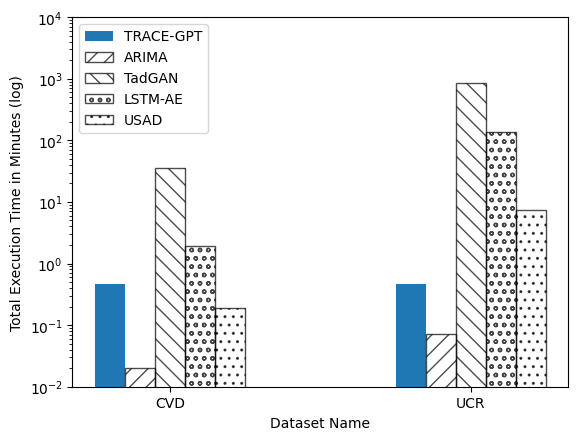} {\textbf{
Comparison of the total execution time in minutes for each dataset. The total execution time consists of training and test time.
}\label{fig:time}}

\textbf{As Table~\ref{tab:baselines}, TRACE-GPT outperforms baseline models based on average of both accuracy and F1 score across all the datasets.} 
Our model showed better performance while maintaining faster runtime compared to most models, as shown in Fig.~\ref{fig:time}. The runtime was longer than that of the traditional statistical model, ARIMA. However, given our model's better performance compared to ARIMA, the increase in runtime can be considered as a worthwhile cost for improved performance. In the case of USAD, it was faster than our model for small datasets, but it suffered from significant increases in runtime as the dataset size grew. When compared to LSTM-AE, our model outperformed it, likely due to the superior performance of Transformers over LSTM in general, attributed to parallel computations \cite{vaswani2017attention}. TadGAN had the least competitive time performance, consistent with findings from other studies \cite{wong2022aer}. Lastly, our model's computation at the sequence level for each wafer, aligning with the characteristics of semiconductor manufacturing data, is also considered a positive factor contributing to its performance enhancement.

\begin{figure}[t!]
\includegraphics[scale=0.55]{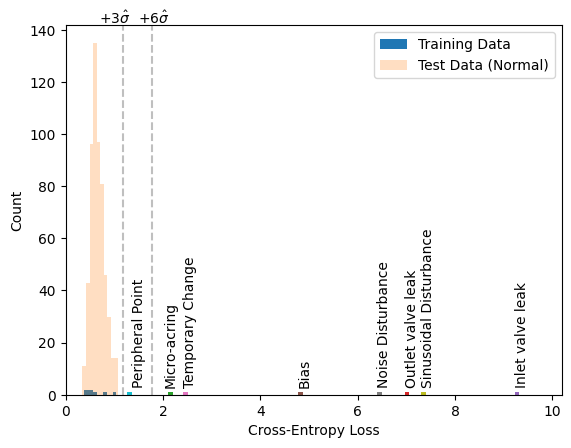}
\caption{
Distribution of the cross-entropy loss ($\mathcal{L}$) on the CVD dataset. 
This histogram shows that test loss has converged, even with the small number of training data.
The losses from faults are higher than those from normal sequences. 
Unlike the UCR dataset, since all fault types are clearly identified, it is possible to compare the loss among different fault types.
Peripheral point was the most challenging fault type to classify.
}\label{fig:cvd_hist}
\end{figure}

\begin{figure*}[t!]
 \centering
 \subfigure[CVD Anomaly Scores] {\includegraphics[scale=0.059]{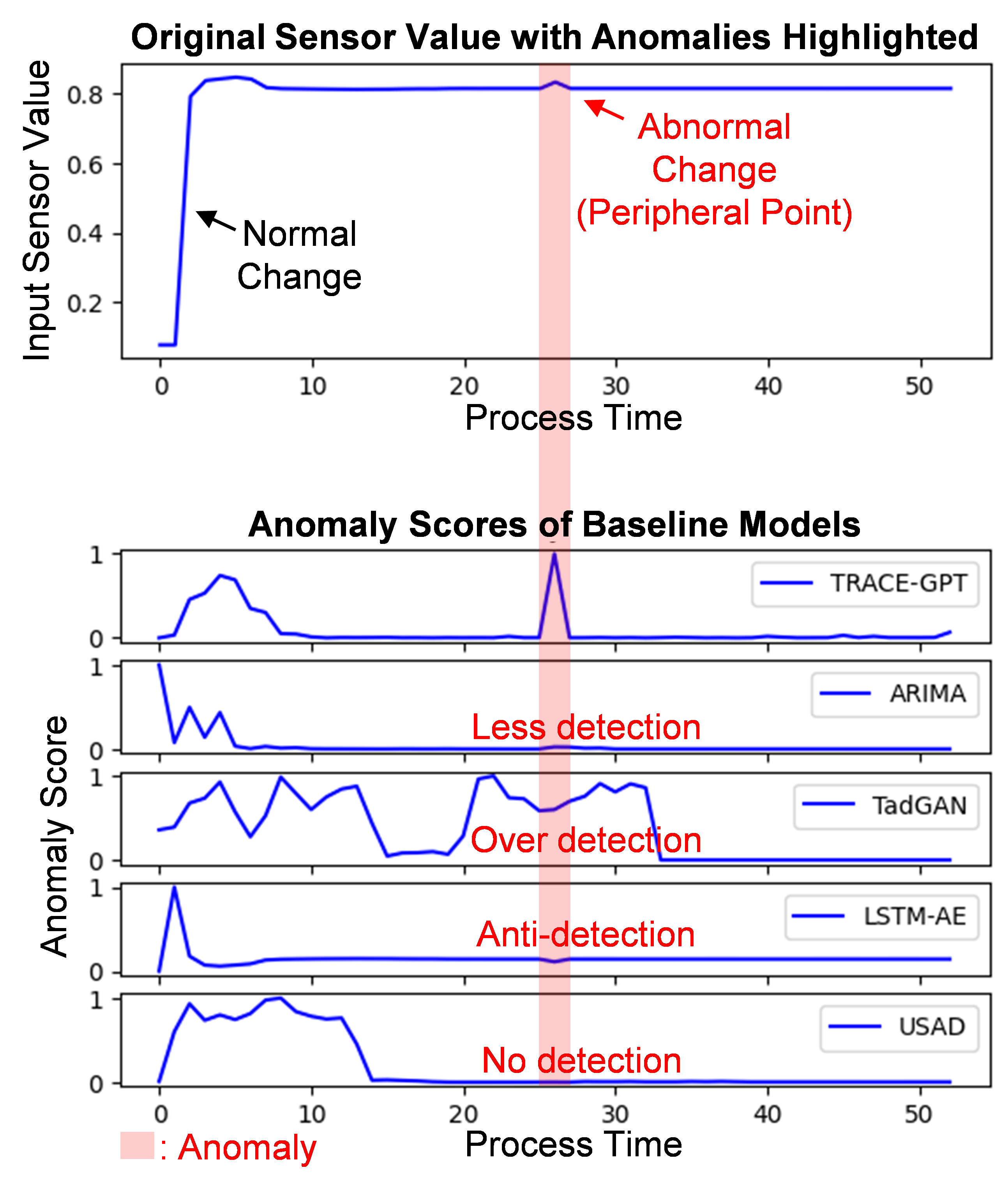}}\hfill
 \subfigure[UCR Anomaly Scores] {\includegraphics[scale=0.059]{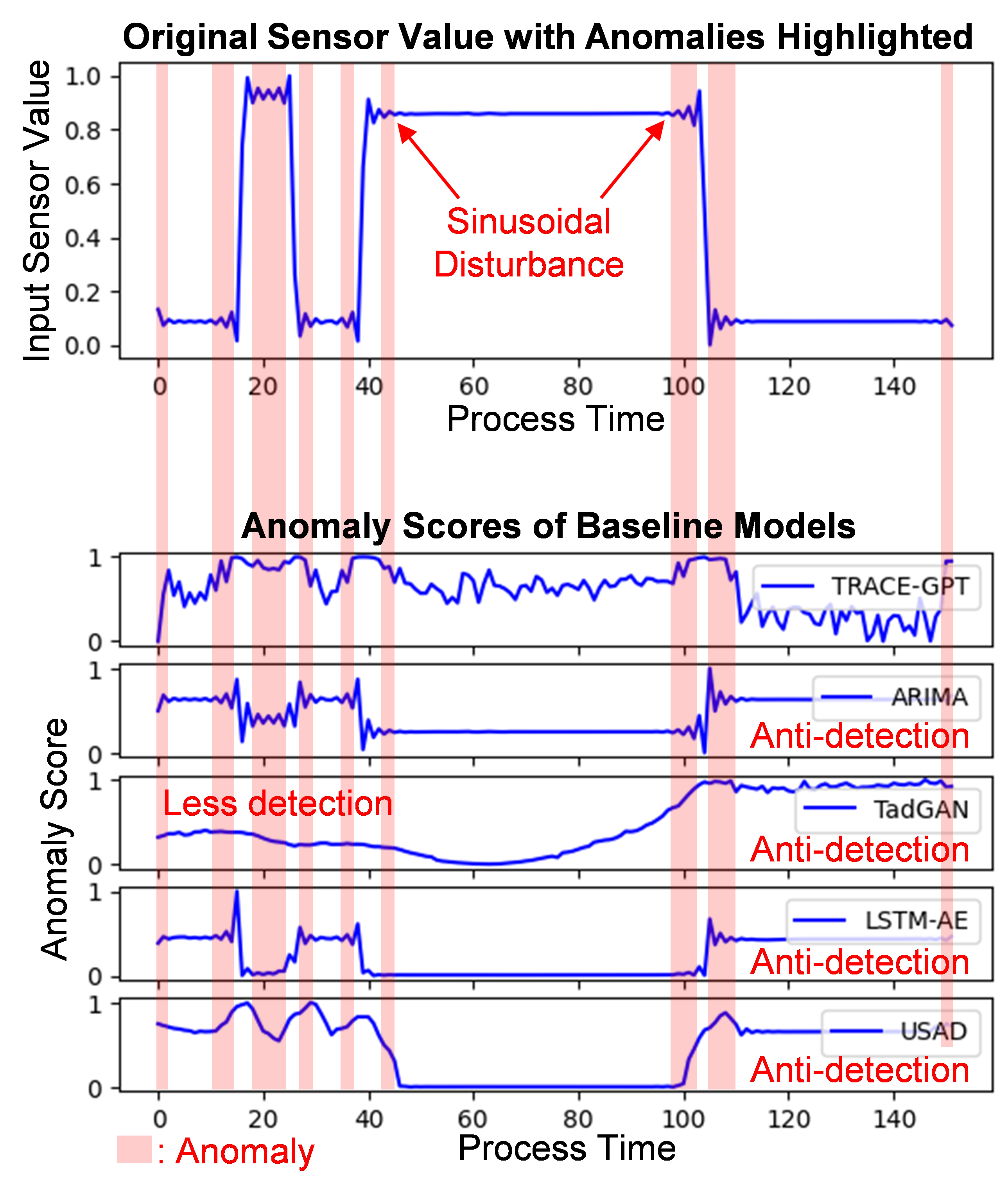}}\hfill
 \subfigure[CVD Prediction by TRACE-GPT] {\includegraphics[scale=0.059]{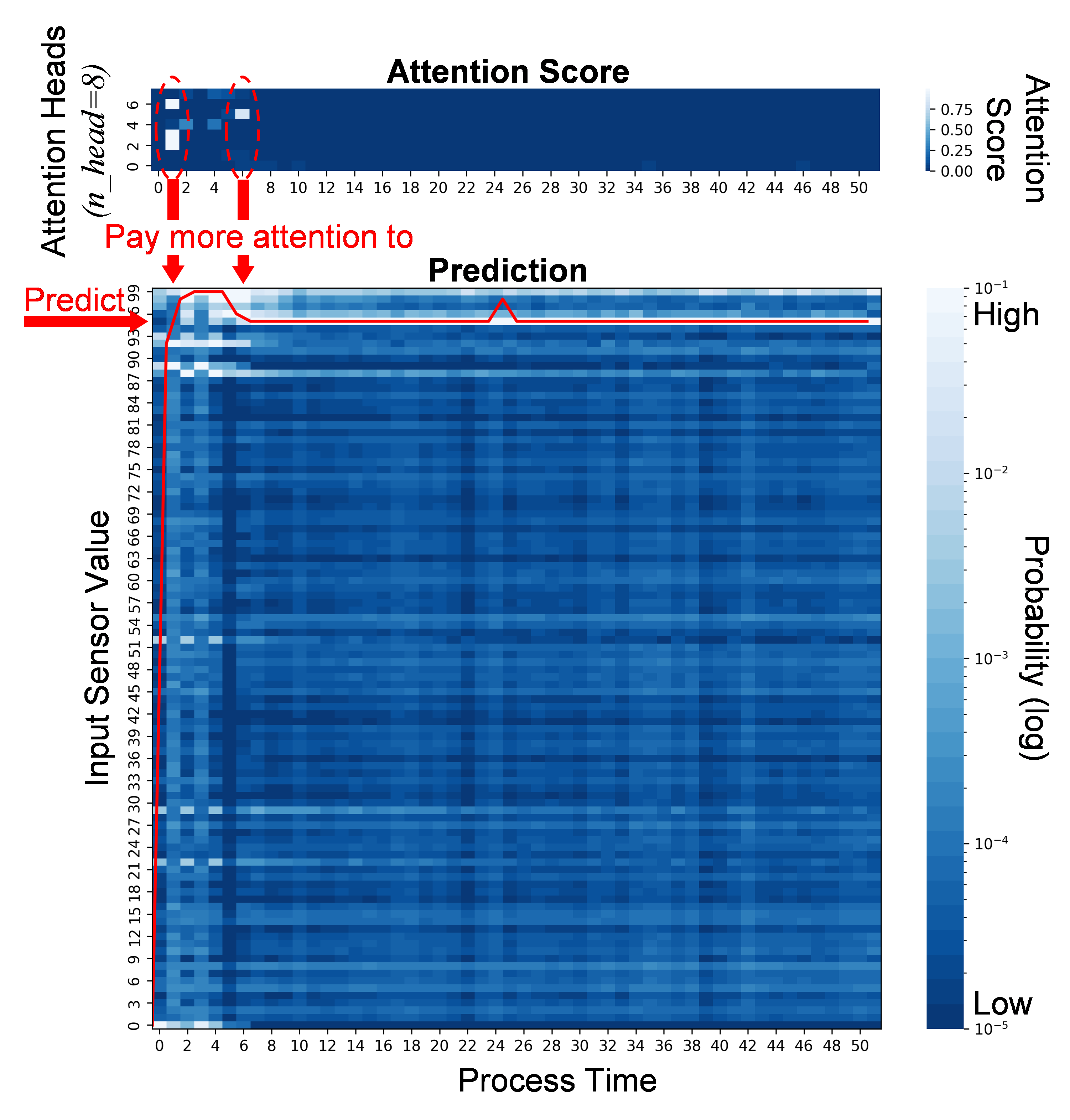}}\hfill
 \subfigure[UCR Prediction by TRACE-GPT] {\includegraphics[scale=0.059]{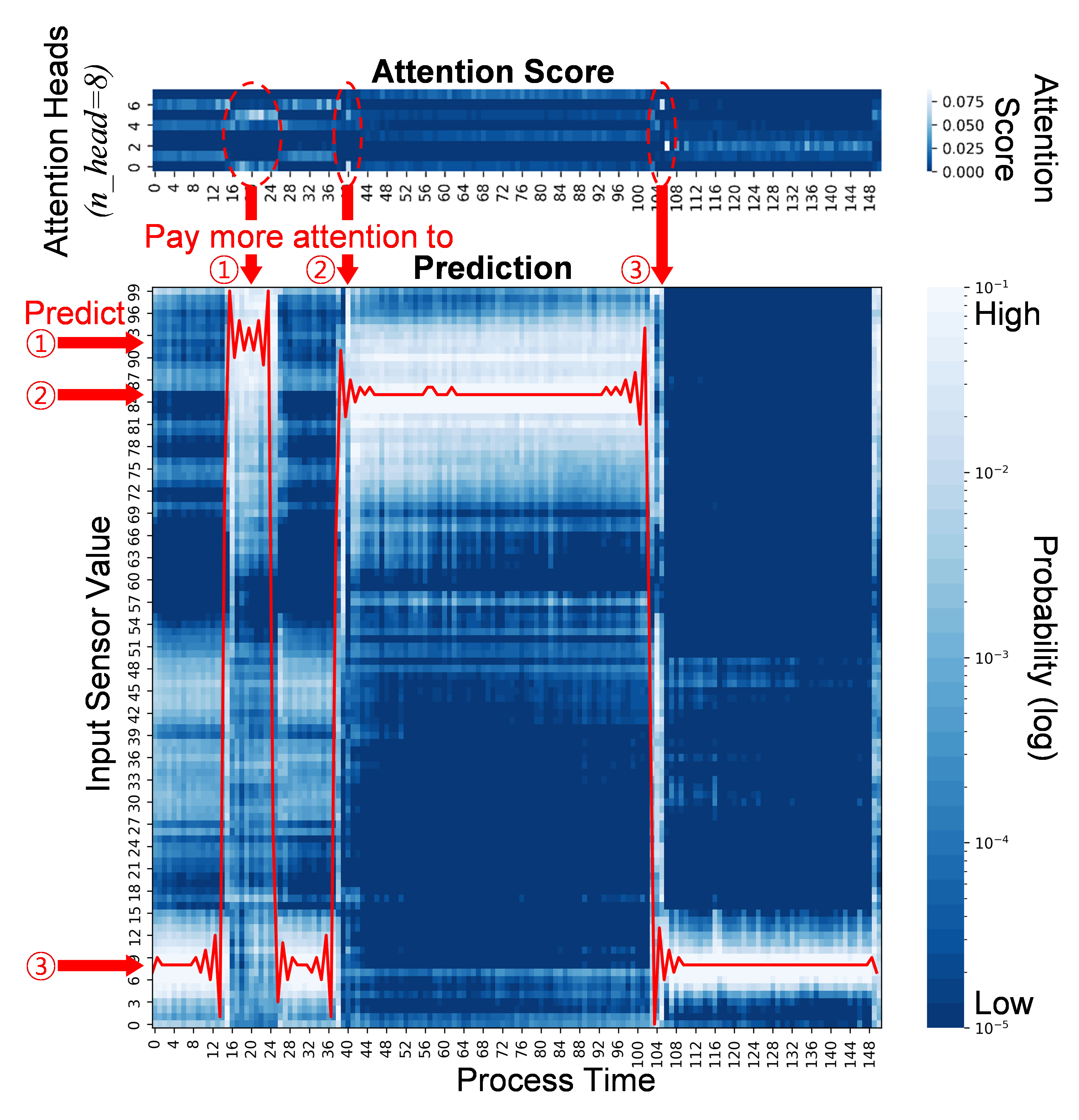}}
 \caption{
Figures (a) and (b) provide examples of how our model and baseline models computed anomaly scores over time for each dataset's given raw data. The ground truth for anomalies (highlighted in red) was confirmed by domain experts in the semiconductor manufacturing process. High anomaly scores should appear during periods of anomalies, while remaining relatively low for the rest.
Figures (c) and (d) represent visualizations of our TRACE-GPT model for the corresponding data. As our model utilizes attention mechanisms, it shows how the eight attention heads learned weights and, based on this, how the model predicts sensor values, visualized through blue heatmap on the background. The red line chart corresponds to the original raw data for the same original time-series as in (a) and (b). 
The sensor values and anomaly scores have been normalized between 0 and 1.
 }
 \label{fig:vis_all}
\end{figure*}

For the CVD dataset, our model successfully distinguishes abnormal wafers from normal wafers without any false classifications. As shown in Fig.~\ref{fig:cvd_hist}, the minimum loss of abnormal wafers exceeds more than three times the estimated standard deviation ($\hat{\sigma}$) of only nine training data points. 
In other words, even with a very limited number of samples, setting the threshold at 3-sigma can yield good performance on a large amount of test data. The results shows that TRACE-GPT can detect time-series anomaly without supervised training. 
Models like ARIMA exhibited limitations in detection performance due to increased noise in normal segments compared to anomaly scores at peripheral points.
In the case of TadGAN, it responded well to peripheral points and demonstrated good detection performance on the CVD dataset. However, due to the reconstruction method's inherent blurring effect, anomaly scores escalated in normal regions, leading to over-detection.
Models like LSTM-AE sometimes yielded lower anomaly scores for timestamps with anomalies when the overall reconstruction shifted. 
Due to the phenomenon where the anomaly score appears lower in regions where it should be detected as high, we have termed this phenomenon "anti-detection." 
Models like USAD exhibited a lack of significant response to very small fluctuations in single values, indicating their potential inadequacy for unsupervised anomaly detection in the semiconductor manufacturing domain.
Across all models, due to the unsupervised nature without prior knowledge, anomaly scores for transition segment that can occur normally in semiconductor manufacturing processes were observed to be higher (See Fig.~\ref{fig:vis_all} (a)).

For the UCR dataset, TRACE-GPT outperformed all the unsupervised baseline methods by having the largest area under the ROC curve (AUC) with 0.9846 as shown in Fig.~\ref{fig:roc}. Comparing to the F1 score of the supervised state-of-the-art result \cite{guillaume2022random}, the F1 score of the TRACE-GPT at Equal Error Rate (EER) is only 0.026 below the supervised baseline. 
As depicted in Fig.~\ref{fig:vis_all} (c), for example, when considering the common sinusoidal disturbance type prevalent in semiconductor manufacturing processes \cite{mellah2022semiconductor}, our model TRACE-GPT exhibited a phenomenon in which the anomaly score was relatively high in the stable interval without anomalies, between time stamps 50 and 90. 
Nonetheless, the scores remained lower compared to the other anomalous points, and consequently, this did not significantly impact the overall detection performance. 
Traditional models like ARIMA demonstrated relatively effective detection of sinusoidal disturbances and showed good performance. 
However, TadGAN did not perform well in detecting anomalies in data from semiconductor manufacturing processes. 
In the case of TadGAN, it had demonstrated good performance on datasets with existing periodic characteristics \cite{geiger2020tadgan}. 
Hence, it is inferred that TadGAN might be vulnerable to datasets with irregular periods or mixed normal shapes, such as the UCR dataset. 
This phenomenon is also applicable to the rest of the baseline models, as even though they successfully assigned high anomaly scores before time stamp 50, a tendency for anti-detection after 110 substantially increased false positives during the classification stage (See Fig.~\ref{fig:vis_all} (b)).

As data visualization plots give a good data explainability \cite{dwivedi2023explainable}, we visualized attentions scores at each attention head and timestamp as shown in Fig.~\ref{fig:vis_all} (c) and (d). To predict a sequence, our model could learn to assign greater weight (pay more attention) to the sensor value that indicates normality within the sequence. If there is a specific abnormal shape, such as sinusoidal disturbance, broader attention scores are assigned to normalize the noise within the abnormality.

\subsection{Ablation Study}
\textbf{The combination of all three components is essential to achieve the highest accuracy and F1 score.} As shown in Table~\ref{tab:ablation}, our experiment demonstrated that three components of our model, namely Positional Embedding (PE), Temporal Convolutional Networks (TCN), and the Transformer, all contribute to detecting anomalies through unsupervised learning.

\textbf{The Transformer architecture is the key component of our model}, which demonstrates that the Transformer can be successfully used in numerical time-series sensor data in the semiconductor manufacturing domain.

\textbf{Both TCN and the Transformer effectively learn time-series features from small-sized training data.} Particularly, Precision in the CVD dataset is significantly improved by both TCN and the Transformer.

\textbf{With mixed types of normality, the Transformer excels in predicting the next sensor value.} Especially, the Transformer significantly enhances Recall in the UCR dataset.

\subsection{Limitations and Discussion}

We compared our model, TRACE-GPT, to well-known baseline models. Even though the performance metrics show progress in solving challenges within real-world datasets of the semiconductor manufacturing industry, there is still room for improvement in our model. 
For instance, TRACE-GPT is a univariate model. As it uses TCN for embedding, multivariate settings can be easily implemented by adjusting the number of input channels in the TCN layer. With a suitable multivariate datasets, this model can be tested as multivariate model in the future. Another improvement involves sequence length. In this study, since all sequences had the same length, there was no need to consider about variable-length \cite{kim2019fault}. The advantage of applying the Transformer model, which is widely used in the NLP domain, to problems with different sequence lengths is that it can be processed by assigning a padding token after the end of sequence token \cite{vaswani2017attention}. With appropriate dataset, we would like to apply our model to variable-length sensor data in future. Lastly, by architecture, anomaly in the first sensor value in the sequence cannot be detected in current setting. This limitation can be solved by bidirectional prediction \cite{schuster1997bidirectional} in future. Moreover, bidirectional setting might improve the prediction performance in pre-training. Therefore, further research can introduce more universal datasets or modification in architecture.

\section{Conclusion}
\label{sect:conclusion}
In this paper, we present a novel framework, TRACE-GPT, demonstrating how convolutional embedding and the Transformer can be effectively used from anomaly detection to classification in time-series data with unsupervised pre-training in the semiconductor industry. We developed a deep learning model that operates similarly to a human's judgment of anomalies when presented with sensor values that deviate from predictions, by altering the embedding method and utilizing the concept of data resolution in the conventional Transformer architecture. We have shown that the TRACE-GPT performs effectively, even with small or mixed datasets, and in scenarios where abnormal data are either unlabeled or non-existent, which are immediate challenges in the semiconductor manufacturing industry. Especially, we validated our model on open dataset, UCR, to prove that this method outperforms other unsupervised anomaly detection methods. Based on the F1 score at EER, our model almost caught up with the supervised state-of-the-art baseline by only 0.026 with unsupervised learning.

\section*{Acknowledgment}
The authors would like to thank Jaeho Jang, Aekyung Kim, and Junmin Lee for helpful discussions and advice. We would also like to thank Denise Priwisch for her help in reviewing the manuscript.

\bibliographystyle{IEEEtran}
\bibliography{main}

\begin{IEEEbiography}[{\includegraphics[width=1in,height=1.25in,clip,keepaspectratio]{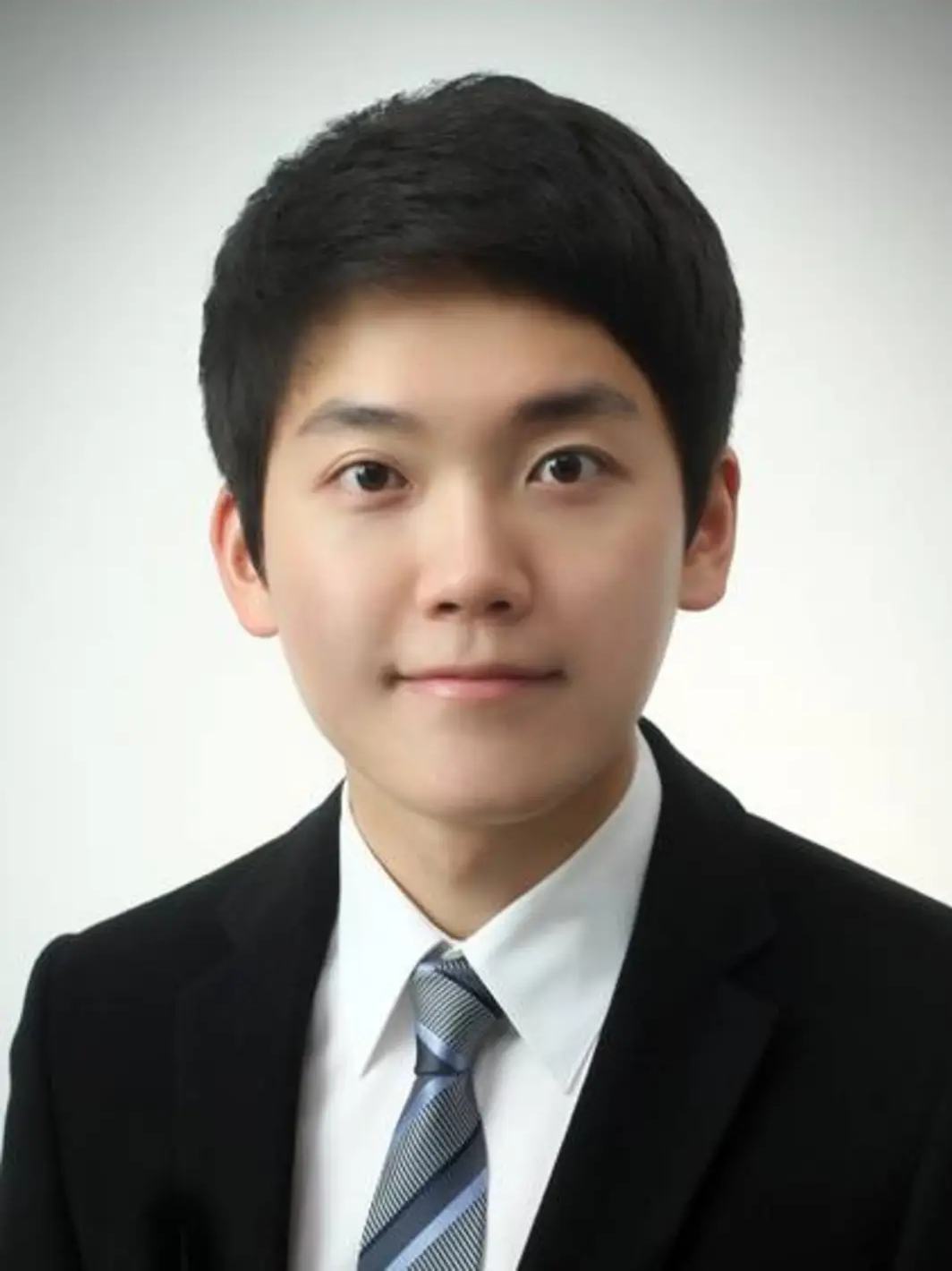}}]{Sewoong Lee} received the B.S. degree in industrial engineering from
the Seoul National University, Seoul, Republic of Korea, in 2015 and the Master of Computer Science degree from the University of Illinois at Urbana-Champaign (UIUC), Urbana, IL, USA, in 2023.

From 2016, he has been a Software Engineer at Mechatronics Research, Samsung Electronics.
His current research interests include data science for the semiconductor manufacturing, time-series anomaly detection, and the development and deployment of novel deep neural networks models. 
\end{IEEEbiography}

\begin{IEEEbiography}[{\includegraphics[width=1in,height=1.25in,clip,keepaspectratio]{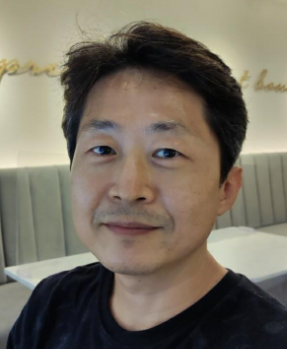}}]{JinKyou Choi} received the M.S. degree in electrical engineering from
the Korea University, Seoul, Republic of Korea, in 2013.

From 2002, he has been a Software Engineer at Mechatronics Research, Samsung Electronics.
His current research interests include data science for the semiconductor manufacturing. 
\end{IEEEbiography}

\begin{IEEEbiography}[{\includegraphics[width=1in,height=1.25in,clip,keepaspectratio]{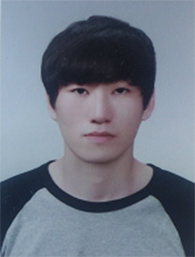}}]{Min Su Kim} received the B.S., M.S., and Ph.D. degrees in electrical engineering from the Pohang University of Science and Technology (POSTECH), Pohang, Republic of Korea, in 2014, 2016 and 2022, respectively.

From 2022, he has been a Staff Engineer at Mechatronics Research, Samsung Electronics. His research research interests include the data-based fault diagnosis, defect detection, signal processing, and deep neural networks.
\end{IEEEbiography}

\EOD

\end{document}